\documentclass{article}

\usepackage{microtype}
\usepackage{graphicx}
\usepackage{subfigure}
\usepackage{booktabs} 

\usepackage{hyperref}

\usepackage[accepted]{icml2021}

\usepackage{times}
\usepackage{epsfig}
\usepackage{amsmath}
\usepackage{amssymb}
\usepackage{comment}
\newcommand{\nop}[1]{}

\icmltitlerunning{On the Difficulty of Generalizing Reinforcement Learning Framework for
Combinatorial Optimization}

\begin{document}

\twocolumn[
\icmltitle{On the Difficulty of Generalizing Reinforcement Learning Framework for \\
           Combinatorial Optimization}



\icmlsetsymbol{equal}{*}

\begin{icmlauthorlist}
\icmlauthor{Mostafa Pashazadeh}{equal,to}
\icmlauthor{Kui Wu}{equal,to}
\end{icmlauthorlist}

\icmlaffiliation{to}{Department of Computer Science, University of Victoria, Victoria, Canada}

\icmlcorrespondingauthor{Mostafa Pashazadeh}{mostafapashazadeh@uvic.ca}
\icmlcorrespondingauthor{Kui Wu}{wkui@uvic.ca}

\icmlkeywords{Machine Learning, ICML}

\vskip 0.3in
]



\printAffiliationsAndNotice{\icmlEqualContribution} 

\begin{abstract}
Combinatorial optimization problems (COPs) on the graph with real-life applications are canonical challenges in Computer Science. The  difficulty of finding quality  labels for  problem  instances  holds  back  leveraging  supervised  learning  across  combinatorial problems.  Reinforcement learning (RL) algorithms have recently been adopted to  solve  this  challenge  automatically.  The  underlying  principle  of  this  approach  is to  deploy  a  graph neural network (GNN) for  encoding  both  the  local  information  of  the nodes and the graph-structured data in order to capture the current state of the environment.   Then, it is followed by the actor to  learn  the problem-specific heuristics on its own and make an informed decision at each state for finally reaching a good solution. Recent studies on this subject mainly focus on a family of combinatorial problems on the graph, such as the travel salesman problem, where the proposed model aims to find an ordering of vertices that optimizes a given objective function. We use the security-aware  phone  clone  allocation  in  the  cloud as a classical quadratic assignment problem (QAP) to investigate whether or not deep RL-based model is generally applicable to solve other classes of such hard problems. Extensive empirical evaluation shows that existing RL-based model may not generalize to QAP.
\end{abstract}

\section{Introduction}

There  is  a  steady  stream  of  approximation algorithms  for  certain classes  of  discrete  optimization  such  as  shortest  paths and  circulation problems. Generally speaking,  each  such  a  problem  has  unique  subtleties  and  constraints  that  prevent people from using a renowned optimization solver for a family of hard problems such as  the  Travel  Salesman  Problem  (TSP)  to  address  all  COPs. This  issue  demands devising  methods  and  examining  heuristics  specific  to  each  individual  COP. Recently, reinforcement learning (RL) algorithms have been successfully applied in solving hard problems. \cite{khalil2017learning, ma2019combinatorial, cappart2020combining} used RL to solve several hard COPs, including Minimum Vertex Cover (MVC), Maximum Cut (MAXCUT),  Vehicle Routing Problem (VRP), Travel Salesman Problem (TSP), and constrained TSP. Experimental results show that this approach can achieve a promising approximation ratio (e.g., the ratio between RL’s tour length and the optimal tour length in TSP). RL-based methods attack COPs through a twofold architecture composed of encoder and decoder. Encoder trains a graph neural network (GNN) to learn the graph representation. The graph structured data is later used to instruct the decoder, made of neural network (NN), to take an astute action. The entire framework is then trained with an RL algorithm to search for a quality policy. Although solutions cannot be proven to be optimal, they get better when the problem’s combinatorial space is further explored and more inputs from problem instances are used to train the agent. To this end, RL algorithms manage to address diverse challenges, including quality solutions and response time, and they are partially successful in dealing with scalability issues.

We are interested in whether or not RL is generally applicable to some typical COPs faced in computer communications and networks. Our important finding is that RL-based approach, at least in its current incarnation, may not generalize successfully to solve quadratic assignment problems (QAPs) \textit{where incrementally constructing an order of nodes is not an inherent requirement}. Furthermore, \textit{the non-trivial objective function} of QAPs may further amplify the difficulty. This insight is a valuable note for researchers trying to promote RL-based solutions in COPs. While it is difficult to provide a theoretical proof on this claim, we offer empirical explanation by carefully analyzing the existing breakthrough in graph encoding techniques and successful RL algorithms on this matter. We also discuss some directions for future research.

\subsection{Statement of the Problem}

Using the security-aware phone clone provisioning problem~\cite{vaezpour2014swap} as the motivating example, we investigate whether or not RL brings benefits and outperforms traditional optimization solutions. The idea of phone clones~\cite{liu2013gearing} is to build software clones of smartphones on the cloud that allow end users to upload resource-consuming tasks and backup data to the cloud. Since the software clone of a smart phone is generally a virtual machine allocated in a server, we need to solve the so-termed phone clone allocation problem, i.e., allocating phone clones on servers under the security and servers' capacity constraints. In this paper, we use server and host interchangeably. 

In practice, a phone clone might hack into others on the same host via covert channel. Hence, it is better to co-locate a phone to those closely connected with the phone rather than the strangers, i.e., assigning phone clones of friends on the same server where ``friendship" of two phones is determined by the communication history between them. However, due to the large number of end-users and the limited number of hosts, it is challenging to perfectly group the connected phone clones and isolate them from strangers. This bottleneck brings up the problem of security-aware provisioning of the phone clones~\cite{vaezpour2014swap}. Although phone clone allocation is identified as a hard problem, its nuance and circumstances differ from the class of COPs mentioned above. In essence, phone clone allocation problem is considered to be a class of quadratic assignment problems (QAPs) called constrained scheduling problems with interaction cost that additionally demand the feasible solution to stay within the constraint  of hosts’ capacities. In the following we introduce a system model that provides a mathematical perspective of the phone clones' security-aware allocation in the cloud. This view helps to recognize its nuance and inherent complexity in contrast with hard problems previously attacked by related work as we go through recent architectures on this matter.

We represent the communication history among mobile users with a weighted communication graph. A smaller weight implies less communication between the endpoints and a higher risk of attacks. We assume the system has $m$ phone clones and $n$ hosts in the cloud. We represent the communication graph with an adjacency matrix $W = [w_{ij}]_{m\times m}$, where $w_{ij}$ is a real value in $[0,1]$ that models the tie between phone clones $i$ and $j$. We denote the phone clone allocation matrix with $X = [x_{ij}]_{m\times n}$, where $x_{ij} = 1$ indicates that phone clone $i$ is allocated to host $j$ and $x_{ij} = 0$ otherwise. Given the adjacency matrix $W$ and the allocation matrix $X$, the potential risk would be formulated as $\Upsilon = \frac{1}{2}tr(X^T\Bar{W}X)$ where $\Bar{W} = [\Bar{w}_{ij}]_{m\times m} = [1-w_{ij}]_{m\times m}$ denotes the complementary adjacency matrix and $\Bar{w}_{ij}$ denotes the potential risk between phone clones $i$ and $j$. $tr(\cdot)$ denotes the trace of a matrix. For security-aware provisioning that keeps to the capacity constraints of hosts, we need to solve the following discrete optimization problem to minimize the risk of a phone clone allocation scheme.
\begin{equation}
\begin{aligned}
    & \underset{X}{min} \hspace{4pt} \frac{1}{2}tr(X^T\bar{W}X) & \\
    s.t. & & \\
    & \sum_{j=1}^{n} x_{ij} = 1, & \text{for } i = 1,2, \dots m \\
    & \sum_{i=1}^{m} x_{ij} \leq c_j, & \text{for } j = 1,2, \dots n 
    \label{eq:opt-prob}
\end{aligned}
\end{equation}
where $c_j$ is the capacity of $j$-th host w.r.t. the maximum number of phone clones that it can host.
\vspace{5mm}

\section{Related Work and Major Differences in Problem Setting}

Bello et al.~\cite{bello2016neural} use a pointer network architecture to solve (2D Euclidean) TSP and KnapSack problems. Given a set of $m$ cities $s = \{x_i\}_{i=1}^m$ where each $x_i \in R^2$ denotes the 2D coordinates of the $i_{th}$ city, they adopt a graph level representation with recurrent neural network (RNN) to read and encode the inputs into a context vector. The decoder \nop{that is made of RNN and an attention function,} takes the context vector and calculates a distribution over the next city to be visited. Decoding proceeds sequentially, that is, once the next city is selected, it is fed to the next decoder step as input. 

There are three inherent differences between TSP and our problem setting that make it nonsensical to apply this model to the phone clone allocation problem. 
\begin{itemize}
  \item The underlying network graph in~\cite{bello2016neural} is assumed to be complete, and graph topology is not incorporated in the encoder whereas in our case we model the communication graph among phone clones with an adjacency matrix.
  \item Inputs in our problem setting would be current allocation recorded in a one-hot vector indicating the host to which a phone clone is assigned. The decoder looks over the node embeddings regardless of the order of the inputs to approximate the action values. As a result, a good encoder should be invariant to the permutation of input vectors so that changing the order of any two input vectors does not affect the model.
  \item The solution (tour in TSP) is built up incrementally which allows the decoder to use a helper function to mask the cities that were already visited. In other words, the masking procedure narrows the focus of the investigation and conducts the decoder to find a better solution. Besides, the mask function rules out the impact of the nodes that have already been touched from the context vector which is passed to the decoder as the surrogate for the environment. This property helps the decoder to discriminate through subsequent iterations as the training episode moves forward. Nonetheless, the mask function is irrelevant to the circumstances of our problem because incrementally constructing an order of nodes is not an inherent requirement. In our problem setting, the counterpart of figuring out the next node (city) to be added to the partial solution (partial tour in TSP) is to decide the host to which we will assign the next phone. In this case, it is not a straightforward task to manually put masking on the hosts' pool or limit the reallocation of some phones. Otherwise, irrational interference, which merely reduces the feasible domain of actions, could push the decoder further toward a poor policy.
\end{itemize}

\cite{kool2018attention} and \cite{vaswani2017attention} push forward the idea of~\cite{bello2016neural} and employ graph attention network that incorporates both the graph topology and input features as opposed to the previous architecture that employed a graph-agnostic sequence to sequence mapping. Each layer of the encoder is roughly a weighted message-passing mechanism between the nodes of graph, which allows a node to receive different types of messages from other nodes. \nop{Then, a combination of these attention vectors would be adopted to constitute the new embedding for the next layer.} The decoder sequentially determines the next node to be visited, one node per iteration. At each time step, the decoder takes the node embeddings and the partial solution (tour) generated until then to create a context node. Then using a single head attention mechanism, the decoder computes the attention weights between the context node and node embeddings that have not been visited yet. These attention weights are then considered to be the distribution over the next node that the decoder adds to the tour. The loss function is defined as the negative tour length, and policy gradient algorithm is used to train the parameters. The above message passing method seems to be a good fit for a class of hard COPs such as TSP, VRP, and OR. Experimental results also show that this model brings benefits over the pointer network in~\cite{bello2016neural}. 

There are, however, some barriers that hinder this method in tackling our problem. 
\begin{itemize}
    \item Encoder incurs overly extra computation that leaves an adverse impact on the performance when it comes to problems of larger size or other variants of hard optimization problems. Experimental results show that this model cannot scale up to solve larger problems, which endorses our claim.
    \item It is not a promising option for weighted graphs as it relies on queries and keys to realize weighted message passing rather than graph's weight matrix itself. \nop{Indeed, we consider applying graph attention network as an alternative solution to a variant of our problem with unweighted graph to asses the competency of this approach in regard to the new problem setting.}
    \item As mentioned above, the essence of the phone clone allocation problem bars the RL-based solution to count on a helper function for masking. \nop{In our case, the optimum policy should be learned by RL framework itself through exploiting the problem's structure.}
\end{itemize}

\cite{ma2019combinatorial} and \cite{khalil2017learning} introduced a variant of message passing methodology which effectively reflects the graph-structured data from environment. \cite{ma2019combinatorial} focused on constrained TSP and \cite{khalil2017learning} addressed TSP, MVC and MAXCUT problems. To update the node embedding at each layer, the encoder takes into account the neighbours of node and the weight of the edge between nodes. Since this model complies with the characteristics of our problem setting, we develop from that to solve the more complex problem of phone clone allocations. Once node embeddings are computed at the last layer, the decoder takes them as inputs and sequentially adds the next node to the partial solution (tour in TSP) in the order determined by its learned policy. The decoder in~\cite{khalil2017learning} uses a fitted Q-learning technique to learn a parameterized policy that aims to optimize the objective function of the problem instance (minimizing the tour length in TSP). The main advantage of Q-learning algorithm is that it is mindful of delayed reward, which makes it a fitted approach for training our problem as well. 

A drawback of this method, similar to previous works, is that it relies on a masking function to reduce the search space. As opposed to previous works that incrementally construct the final solution starting from a partial solution (tour), the phone clone provisioning improves on current provisioning of the clones by reallocating the phone clones in the cloud. 

\section{Framework}

\begin{figure}[h]
\centering
\includegraphics[width=1.0\linewidth]{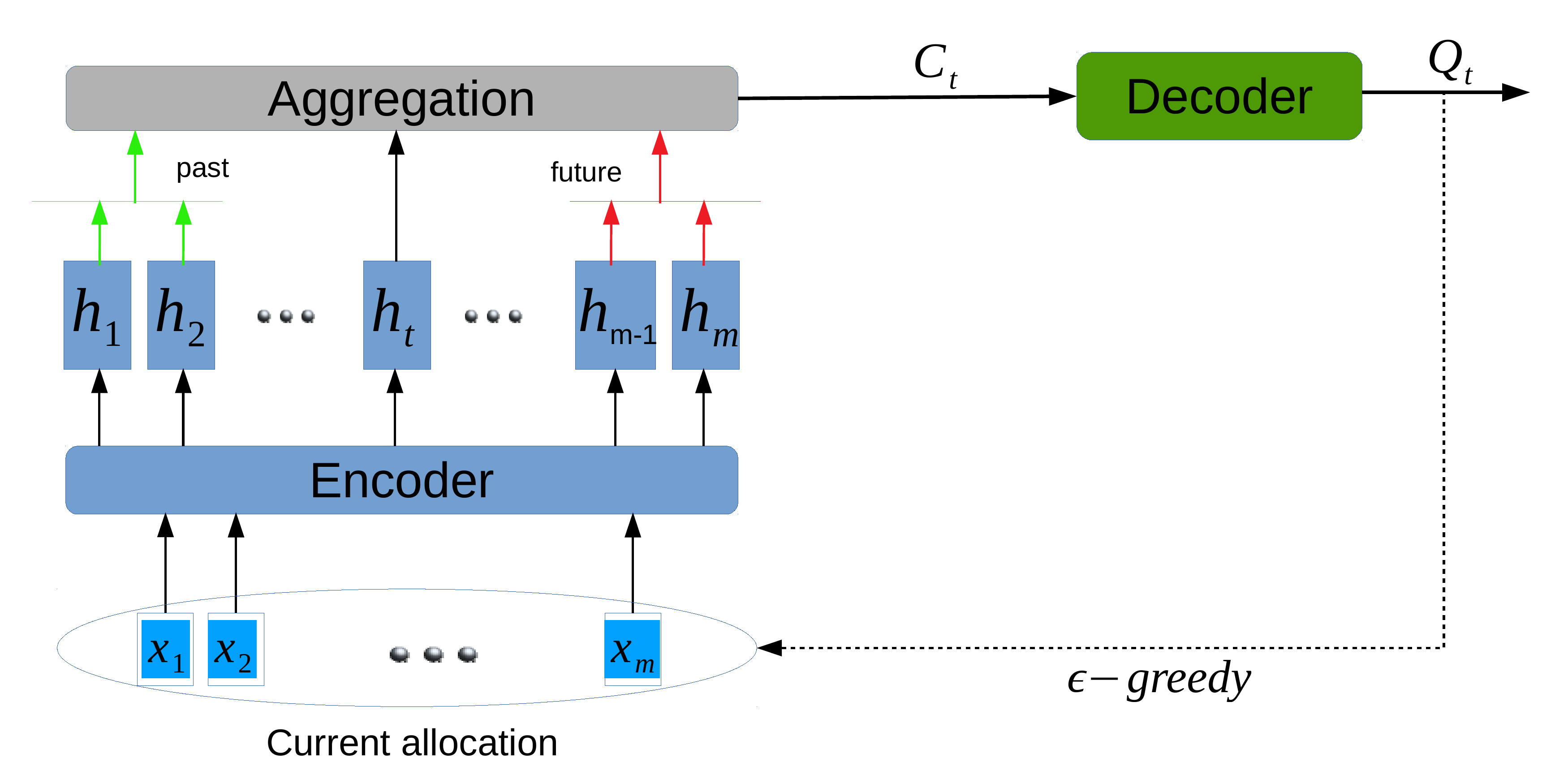}
\caption{Q-learning model architecture.}
\label{fig:Qmodel}
\end{figure}

Since the same high-level design of previous works cannot be directly applied to the phone clone allocation problem, we need a fitted model to tackle this issue. For this, we first build an architecture with a GNN capable of concisely embedding both current provisioning of the clones (solution-specific features) and the underlying communication graph (problem-specific features). The GNN is followed by an aggregation layer that looks over these embeddings and provides meaningful features of the current state to the decoder. Fig.~\ref{fig:Qmodel} shows an overview of the architecture.

The encoder receives the current allocation matrix (solution) as a set of allocation vectors $x_i, \hspace{3pt} i \in \{1, \dots , m\}$. It first maps $x_i \in R^n$ into a higher dimensional vector to produce initial node embedding $h_i^0 \in R^{d_h}$. It builds several attention layers upon initial embedding, each updating the node embeddings repeatedly according to
\begin{equation}
    h_i^{(l+1)} = relu(\theta_3 x_i + \theta_2 \sum_{j=1}^m \bar{w}_{ij} \hspace{2pt} relu(\theta_1 h_j^{(l)} + \mu_1)),
    \label{eq:encoder}
\end{equation}
where $\theta_3 \in R^{d_h \times n}$, $\theta_2, \theta_1 \in R^{d_h \times d_h}$ and $\mu_1 \in R^{d_h}$ are trainable parameters, the $relu$ function introduces nonlinear operation and $\bar{w}_{ij} = 1 - w_{ij}$ is the complementary of the weighted edge between nodes $i$ and $j$. $h_i^{(l)}$ and $h_i^{(l+1)}$ are the inputs and outputs of layer $l$, respectively. To make message passing more powerful, Eq.~(\ref{eq:encoder}) adds a pre-pooling operation represented by parameters $\theta_1$ and $\mu_1$ followed by $relu$ before looking over node embeddings. Also, it incorporates a residual path from initial embeddings \nop{$h_i^{(0)} = \theta_3 x_i$} at each attention layer.

This model can be initialized in any state (not necessarily feasible) and seeks to improve on any proposed solution. Once the final embedding for each node is computed after $L$ recursions, the embeddings are passed to the actor as the environment state. The actor reallocates a phone clone per time step (iteration), and improves on current solution. At each iteration $t \in \{1, \dots , m\}$, the actor first aggregates node embeddings that were reallocated before, with respect to communication history between the phone at hand and reallocated phones. This is done through the aggregation layer on top of the encoder, and the outcome is termed left context vector $c_t^{left} = \sum_{t'=1}^{t-1} \bar{w}_{tt'} h_{t'}^{(L)}$, where $h_{t'}^L$ is the embedding of node $t'$ from last attention layer. It then aggregates node embeddings that are going to be reallocated in the rest of the episode, and outcome is termed right context vector $c_t^{right} = \sum_{t'=t}^m \bar{w}_{tt'} h_{t'}^{(L)}$. Breaking down the context vector into $c_t^{left}$ and $c_t^{right}$ brings the benefit to allow the decoder to figure out more reliable action values at each iteration. Decoder in the phase receives these context vectors as input. It then processes them through separate channels to acquire a better view of the environment state. We use neural network as parameterized Q-function to approximate the state-action values as follows.
\begin{equation}
\begin{aligned}
    & Q(c_t; \Theta) = \theta_6 relu(c_t), 
    & c_t = [\theta_5 c_t^{left}, \theta_4 c_t^{right}], 
    \label{eq:qfunc}
\end{aligned}
\end{equation}
where $c_t$ is called context vector and $[\cdot , \cdot]$ is the concatenation operation. $\theta_4 , \theta_5 \in R^{d'_h \times d_h}$, and $\theta_6 \in R^{n \times 2d'_h}$ are trainable parameters. $Q(c_t; \Theta)$ is an $n$-dimensional vector that represents the values of allocating current phone to any of hosts depending on a set of $6$ trainable parameters $\Theta = \{\theta_i\}_{i=1}^6$. Accordingly, decoder makes $\epsilon$-greedy decision $a_t$ with respect to the state-action values, $Q(c_t; \Theta)$, following a learned optimal policy $\pi$. $a_t$ here determines the host to which the current phone clone will be allocated. After each decision, the node embeddings get updated based on the new allocation matrix and the underlying communication graph to reflect the change in the problem environment after reallocating the most recent phone.

We deploy k-step Q-learning algorithm to train this model because it considers the delayed reward. It gives the model the ability to allocate the current phone clone to a relatively high risk host while planning to displace some problematic phone clones from this host later in the training episode. The function approximator's parameters get updated at each iteration of an episode through the gradient descent step to minimize the square loss
\begin{equation}
\begin{aligned}
    & loss = (Q^{\pi}(c_t; \Theta)- (r_{t:t+k} + \gamma \hspace{2pt} max \hspace{2pt} \hat{Q}(c_{t+k}; \Theta))^2 & \\
    \label{eq:loss}
\end{aligned}
\end{equation}
where $Q^{\pi}(c_t; \Theta)$ represents action value taken by the $\epsilon$-greedy $\pi$. $\gamma$ is a discount factor. $r_{t:t+k} = \sum_{t'=0}^{k-1} r_{t+t'}$ is the accumulated reward over the next $k$ iterations where reward $r_t$ at iteration $t$ is defined as the cut back in total potential risk and penalty term after new phone clone is reallocated. We impose the penalty term for infeasible solution on the reward function to push the actor toward a feasible solution with respect to the constraints in (\ref{eq:opt-prob}). Given the constraint on hosts' capacities, we set a desired distribution of phones over the hosts and define the penalty term in regard to the Kullback–Leibler (KL) divergence between the actual distribution and the desired distribution. Since the intermediate terms would be canceled out, after $k$ iterations, the outcome amounts to the gap between the last KL term and the first KL term, yielding either positive or negative outcome. In a sense, the penalty term measures how the actual distribution becomes closer to the desired one after the last allocation compared to $k$ steps earlier.

Algorithm~\ref{alg:alg1} concisely explains the training process. The model is trained over multiple episodes, each with a graph instance generated through the stochastic block model~\cite{airoldi2008mixed}, which creates communication graphs with several clusters. To construct the graphs in our experiments, we use \textit{SparseBM} (SBM), which is a python module for handling sparse graphs with clusters given the connection probabilities within and between clusters.
\begin{algorithm}[tb]
   \caption{k-step Q-learning}
   \label{alg:alg1}
\begin{algorithmic}
   \STATE {\bfseries Input:} number of episodes $Z$, and replay batch $B$
   \STATE {\bfseries Output:} $\Theta$
   \STATE Initialize experience replay memory $M$ to capacity $N$
   \STATE Initialize agent network parameters $\Theta$
   \FOR{episode = 1 {\bfseries to} Z}
   \STATE $g \sim$ SampleGraph($G$)
   \STATE $s \sim$ SampleSolution($S$)
   \FOR{i = 1 {\bfseries to} m}
   \STATE Compute the context vector $c_i$
   \STATE $a_i=\left\{\begin{array}{ll} 
                \text{random host} & w.p. \hspace{5pt} \epsilon \\
                argmax \hspace{2pt} Q(c_i;\Theta) & \text{otherwise}\\
                \end{array}
                \right.$
   \IF{$i \geq k$}
   \STATE Add tuple $(c_{i-k}, a_{i-k}, r_{i-k:i}, c_i)$ to $M$
   \ENDIF
   \STATE Sample rand batch from replay memory $B \overset{i.i.d}{\sim} M$
   \STATE Update $\Theta$ by Adam given the average loss over $B$
   \ENDFOR
   \ENDFOR
\end{algorithmic}
\end{algorithm}

\section{Experiments}

\begin{figure}[h]
    \centering
    \includegraphics[width=1\linewidth]{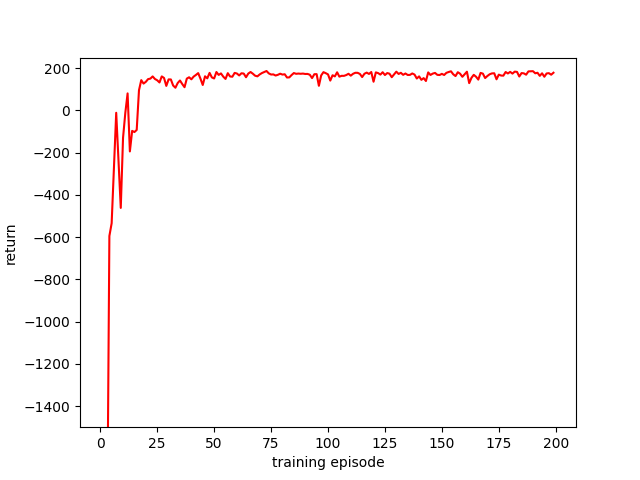}
    \caption{Training progress over time, numb of phones $= 100$.}
    \label{fig:tr-sbm}
\end{figure}

\begin{figure}[h]
    \centering
    \includegraphics[width=1\linewidth]{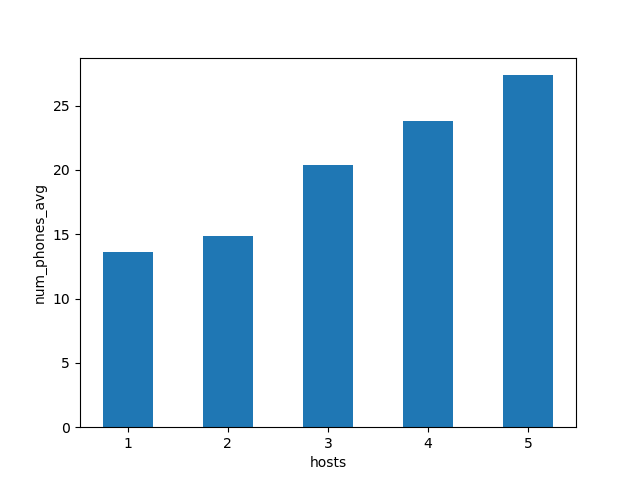}
    \caption{\centering Num of phones allocated to the hosts, numb of phones $= 100$.}
    \label{fig:const-sbm}
\end{figure}

We train the model on communication graphs with $100$ phones allocated to $5$ hosts for the optimization problem~(\ref{eq:opt-prob}). The five hosts are set with different \textit{relative} capacities as $[0.1, 0.1, 0.2, 0.3, 0.3]$. Fig.~\ref{fig:tr-sbm} reports the training progress. We can see that the policy network is gaining more return over training episodes, showing policy improvement during the course of training. The return is the cut back in the potential risk and the penalty term. It is computed at the end of each episode by getting the average over a batch of problem instances. Fig.~\ref{fig:const-sbm} shows the number of phone clones allocated to the hosts. The results fairly keep to the constraints regarding the maximum number of the phones that each host could serve.

\begin{figure}[h]
    \centering
    \includegraphics[width=1\linewidth]{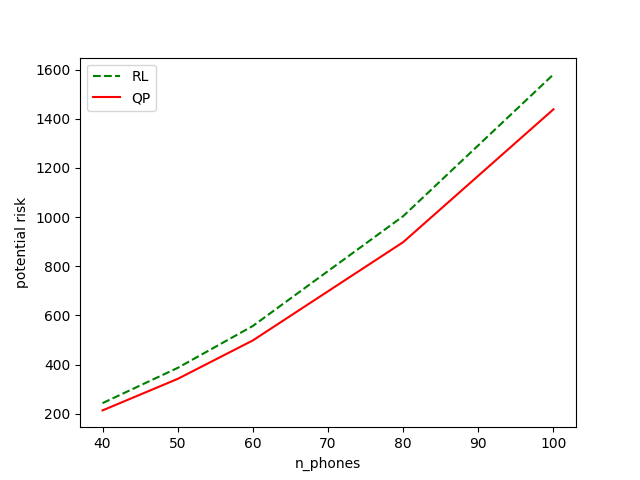}
    \caption{Total potential risk; SBM, number of hosts $= 5$, and relative hosts' capacities are set to $[0.1, 0.1, 0.2, 0.3, 0.3]$.}
    \label{fig:rl-risk-const-sbm}
\end{figure}

\begin{figure}[h]
    \centering
    \includegraphics[width=1\linewidth]{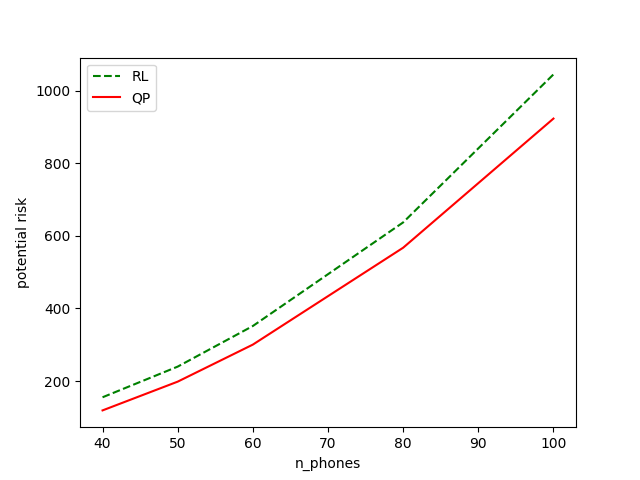}
    \caption{Total potential risk; random graph, number of hosts $= 5$, and relative hosts' capacities are set to $[0.1, 0.1, 0.2, 0.3, 0.3]$.}
    \label{fig:rl-risk-const}
\end{figure}

The trained models are then tested on a range of graph sizes, from $40$ nodes to $100$ nodes. The edge between two arbitrary nodes comes from a uniform distribution over $(0,1)$. For each problem size, we test the model on a batch of graphs and report the averaged result. Moreover, to measure the quality of the results obtained from the RL algorithm, we develop an effective \textit{approximate} solution by exploiting the structure of the problem~(\ref{eq:opt-prob}). We relaxed the phone clone allocation problem to traditional quadratic programming (QP) and then solved it via \textit{trust-constr} method, a general-purpose solver available in \textit{SciPy} library. The performance of this approach is on par with or better than the \textit{approximation} algorithms already employed to solve the phone clone allocation problem~\cite{vaezpour2014swap}. \nop{However, QP solver is a costly approach to our problem setting and is only suitable for solving our problem in small scale.}Figs.~\ref{fig:rl-risk-const-sbm} and~\ref{fig:rl-risk-const} show the potential risk obtained from the RL-based method and that from the QP-based method. As we can see the RL-based method consistently performs poorly than the QP-solvers in different network topologies, indicating that existing RL framework may be hard to generalize for tackling our quadratic assignment problem. \nop{In addition, Fig.~\ref{fig:rl-risk-const} shows that there is also a wide gap between two approaches in the scenario where the graph instances are drawn from uniform distribution in which the edge between two arbitrary nodes comes from a uniform distribution over $(0,1)$.}
\section{Discussion}

Unlike the previous successful reinforcement learning applications to hard problems~\cite{bello2016neural,kool2018attention,vaswani2017attention,khalil2017learning,ma2019combinatorial}, existing RL framework fails to work well in the phone clone allocation problem. In what follows, we highlight the main features of problems where RL is successful, and the lack of such features causes the RL algorithm to under-perform in our problem setting. \textit{First}, incremental approach in previous works allows the decoder to mask the partial solution and narrows the focus of the investigation. Moreover, excluding the partial solution from the encoder's final node embedding makes the context vectors largely distinguishable as the environment state changes from one iteration to another. This quality is boosted when coupled with diverse input embeddings. \nop{As a result, they thoroughly turn over to the decoder the change in the environment after the most recent allocation.} As a result, informative context vectors empower the decoder to figure out the state-action values more accurately. This issue shows the importance of diverse input embeddings and distinctive context vectors. \textit{Second}, in previous problem settings, the policy network learns from a relatively trivial objective function, which is a simple summation over a quantitative feature of the nodes, e.g., the total length of a tour in TSP. However, in the phone allocation problem, the RL algorithm has to minimize an arduous nonlinear cost function through interaction with the environment.

As evidenced by our investigation, due to the inherent complexity of the quadratic assignment problem, building an RL model that can directly produce a satisfactory solution is extremely challenging, if not impossible. Due to the \textit{irreversible} nature of most RL models, the actor cannot revisit and revise its previous decisions. This feature, however, may be highly desirable due to the complexity of the quadratic assignment problem. Thus, a promising future line of work is to enable the actor to reconsider and change its earlier decision. To the best of our knowledge, existing RL models for solving combinatorial optimization problems do not offer an \textit{effective} mechanism to revoke/revise previous decisions. The only work that has adopted a mechanism to address the irreversible nature of the existing methods and revise the previous decisions is~\cite{barrett2019exploratory} by Barrett et al. They worked on the maximum cut problem (not a QAP) and applied this mechanism along with ensemble technique and an intermediate reward procedure to mitigate the problem of locally optimum state. Their work, however, only brought marginal performance improvement over~\cite{khalil2017learning}. Considering the marginal benefit of this method over~\cite{khalil2017learning} and the non-trivial challenge in restructuring this idea to be applicable in our problem (e.g., the complex objective function), we are not optimistic that the RL structure in~\cite{barrett2019exploratory} is powerful enough to solve our problem.

\nop{After $K$ round of message passing via the encoder, a readout function predicts the action value of adding or removing each vertex from the solution subset.} \nop{To apply this idea to our problem setting,}

\section{Conclusion and Future Research}

Using RL to tackle hard COPs is a promising and fantastic research that finds many real-world applications. While we have seen the success of RL in solving several canonical COPs such as TSP, MVC, and MAXCUT, it is natural to investigate whether or not RL can be generalized to solve all COPs. If not, what kind of inherent features in the problem setting renders the success of RL? 

Using the phone clone allocation problem as an example, we show that existing RL framework might be difficult to generalize for solving QAPs. Our investigation also answers the inherent differences between our QAP and the canonical COPs where RL has been capable. While we did not come out of an effective RL solution for the phone clone allocation problem, we believe this type of QAPs poses open challenges for the RL community. The following lines of research deserve future investigation to see if discrete QAPs can be finally tackled by a combination of deep graph network and RL algorithms:
\begin{enumerate}
    \item We might need a twofold policy: $Q_1$, for a scoring mechanism to find the apt phone to reallocate, $Q_2$, for computing the risk of allocating the attained phone to different hosts. Besides, the decision should be feasible with respect to hosts' capacities. 
    \item We need a more sophisticated architecture and rewarding procedures such as hierarchical design. Hierarchical architecture breaks down the complex task into several simple tasks with separate reward functions. Particularly, the lower layer of the hierarchy can be tasked with determining the ``tricky" phone clone regardless of whether it was already reallocated in the ongoing episode. The higher layer can reallocate the phone clones based on the latent data received from the previous layer. To harness this utility, designing reasonable reward functions and strategy for training the model is another future path toward further improving the model.
    \item For the actor to exploit this freedom of going back and revising the former decisions, it is essential to provide the actor with augmented information from the environment. Augmented and rich input vectors also make the states during training distinctively different in the actor's view; resolving the major downside of the untraceable environment states. Useful latent information to incorporate into the input vectors includes the potential risk that the phones currently incur, the potential risk of the phone's host, and steps since the phone clone was last reallocated.
\end{enumerate}

\bibliography{paper_workshop_icml}

\begin{thebibliography}{10}
\providecommand{\natexlab}[1]{#1}
\providecommand{\url}[1]{\texttt{#1}}
\expandafter\ifx\csname urlstyle\endcsname\relax
  \providecommand{\doi}[1]{doi: #1}\else
  \providecommand{\doi}{doi: \begingroup \urlstyle{rm}\Url}\fi

\bibitem[Airoldi et~al.(2008)Airoldi, Blei, Fienberg, and
  Xing]{airoldi2008mixed}
Airoldi, E.~M., Blei, D.~M., Fienberg, S.~E., and Xing, E.~P.
\newblock Mixed membership stochastic blockmodels.
\newblock \emph{Journal of machine learning research}, 9\penalty0
  (Sep):\penalty0 1981--2014, 2008.

\bibitem[Barrett et~al.(2019)Barrett, Clements, Foerster, and
  Lvovsky]{barrett2019exploratory}
Barrett, T.~D., Clements, W.~R., Foerster, J.~N., and Lvovsky, A.~I.
\newblock Exploratory combinatorial optimization with reinforcement learning.
\newblock \emph{arXiv preprint arXiv:1909.04063}, 2019.

\bibitem[Bello et~al.(2016)Bello, Pham, Le, Norouzi, and
  Bengio]{bello2016neural}
Bello, I., Pham, H., Le, Q.~V., Norouzi, M., and Bengio, S.
\newblock Neural combinatorial optimization with reinforcement learning.
\newblock \emph{arXiv preprint arXiv:1611.09940}, 2016.

\bibitem[Cappart et~al.(2020)Cappart, Moisan, Rousseau, Pr{\'e}mont-Schwarz,
  and Cire]{cappart2020combining}
Cappart, Q., Moisan, T., Rousseau, L.-M., Pr{\'e}mont-Schwarz, I., and Cire, A.
\newblock Combining reinforcement learning and constraint programming for
  combinatorial optimization.
\newblock \emph{arXiv preprint arXiv:2006.01610}, 2020.

\bibitem[Khalil et~al.(2017)Khalil, Dai, Zhang, Dilkina, and
  Song]{khalil2017learning}
Khalil, E., Dai, H., Zhang, Y., Dilkina, B., and Song, L.
\newblock Learning combinatorial optimization algorithms over graphs.
\newblock In \emph{Advances in Neural Information Processing Systems}, pp.\
  6348--6358, 2017.

\bibitem[Kool et~al.(2018)Kool, Van~Hoof, and Welling]{kool2018attention}
Kool, W., Van~Hoof, H., and Welling, M.
\newblock Attention, learn to solve routing problems!
\newblock \emph{arXiv preprint arXiv:1803.08475}, 2018.

\bibitem[Liu et~al.(2013)Liu, Shu, Jin, Ding, Yu, Niu, and Li]{liu2013gearing}
Liu, F., Shu, P., Jin, H., Ding, L., Yu, J., Niu, D., and Li, B.
\newblock Gearing resource-poor mobile devices with powerful clouds:
  architectures, challenges, and applications.
\newblock \emph{IEEE Wireless communications}, 20\penalty0 (3):\penalty0
  14--22, 2013.

\bibitem[Ma et~al.(2019)Ma, Ge, He, Thaker, and Drori]{ma2019combinatorial}
Ma, Q., Ge, S., He, D., Thaker, D., and Drori, I.
\newblock Combinatorial optimization by graph pointer networks and hierarchical
  reinforcement learning.
\newblock \emph{arXiv preprint arXiv:1911.04936}, 2019.

\bibitem[Vaezpour et~al.(2014)Vaezpour, Zhang, Wu, Wang, and
  Shoja]{vaezpour2014swap}
Vaezpour, S.~Y., Zhang, R., Wu, K., Wang, J., and Shoja, G.~C.
\newblock Swap: Security aware provisioning and migration of phone clones over
  mobile clouds.
\newblock In \emph{2014 IFIP Networking Conference}, pp.\  1--9. IEEE, 2014.

\bibitem[Vaswani et~al.(2017)Vaswani, Shazeer, Parmar, Uszkoreit, Jones, Gomez,
  Kaiser, and Polosukhin]{vaswani2017attention}
Vaswani, A., Shazeer, N., Parmar, N., Uszkoreit, J., Jones, L., Gomez, A.~N.,
  Kaiser, {\L}., and Polosukhin, I.
\newblock Attention is all you need.
\newblock In \emph{Advances in neural information processing systems}, pp.\
  5998--6008, 2017.

\end{thebibliography}
\bibliographystyle{icml2021}

\end{document}